\documentclass{article}

\usepackage[nonatbib, final]{neurips_2024}
\usepackage{graphicx}
\usepackage{sidecap}

\usepackage[utf8]{inputenc} 
\usepackage[T1]{fontenc}    
\usepackage{hyperref}       
\usepackage{url}            
\usepackage{booktabs}       
\usepackage{multirow}
\usepackage{amsfonts}       
\usepackage{nicefrac}       
\usepackage{microtype}      
\usepackage{xcolor}         
\usepackage{float}                  
\usepackage{subfig}                 
\usepackage{overpic}

\usepackage{caption}

\usepackage{xspace}
\usepackage{mathrsfs}
\usepackage{amsmath}
\usepackage{algorithmic}
\usepackage{algorithm}
\newcommand{\method}{CDC\xspace}
\newcommand{\methodfull}{Causality-Guided Semantic Decoupling and Classification\xspace}
\newcommand{\deshort}{VSD\xspace}
\newcommand{\defull}{Visual-Language Dual Semantic Decoupling\xspace}
\newcommand{\meshort}{DSTC\xspace}
\newcommand{\mefull}{Decoupled Semantic Trusted Classification\xspace}

\usepackage{wrapfig}

\title{Rethinking Misalignment in Vision-Language Model Adaptation from a Causal Perspective}

\author{%
  Yanan Zhang\footnotemark[1]  \\
  University of Chinese Academy of Sciences\\
  Institute of Software Chinese Academy of Sciences\\
  \texttt{yanan2018@iscas.ac.cn}
  \And
  \ Jiangmeng Li\thanks{Equal contributions.} \ ,\ Lixiang Liu\ , \ Wenwen Qiang\thanks{Corresponding author.} \\
  Institute of Software Chinese Academy of Sciences\\
  \texttt{\{jiangmeng2019, lixiang, qiangwenwen\}@iscas.ac.cn}
}

\author{%
  Yanan Zhang\textsuperscript{\rm 1,2,}\footnotemark[1], Jiangmeng Li\textsuperscript{\rm 2,}\thanks{Equal contributions. }, Lixiang Liu\textsuperscript{\rm 1,2}, Wenwen Qiang\textsuperscript{\rm 2,}\thanks{Corresponding author. }\\
  \textsuperscript{\rm 1}University of Chinese Academy of Sciences\\
  \textsuperscript{\rm 2}Institute of Software Chinese Academy of Sciences\\
  \texttt{zhangyanan110199@gmail.com}, \texttt{\{jiangmeng2019, lixiang, qiangwenwen\}@iscas.ac.cn} \\
}

\begin{document}

\maketitle

\begin{abstract}
Foundational Vision-Language models such as CLIP have exhibited impressive generalization in downstream tasks. However, CLIP suffers from a two-level misalignment issue, i.e., task misalignment and data misalignment, when adapting to specific tasks. Soft prompt tuning has mitigated the task misalignment, yet the data misalignment remains a challenge. To analyze the impacts of the data misalignment, we revisit the pre-training and adaptation processes of CLIP and develop a structural causal model. We discover that while we expect to capture task-relevant information for downstream tasks accurately, the task-irrelevant knowledge impacts the prediction results and hampers the modeling of the true relationships between the images and the predicted classes. As task-irrelevant knowledge is unobservable, we leverage the front-door adjustment and propose \methodfull (\method) to mitigate the interference of task-irrelevant knowledge. Specifically, we decouple semantics contained in the data of downstream tasks and perform classification based on each semantic. Furthermore, we employ the Dempster-Shafer evidence theory to evaluate the uncertainty of each prediction generated by diverse semantics. Experiments conducted in multiple different settings have consistently demonstrated the effectiveness of \method.
\end{abstract}
\section{Introduction}
Benefiting from large-scale training data, foundational Vision-Language models such as CLIP~\cite{DBLP:conf/icml/RadfordKHRGASAM21}, ALIGN~\cite{DBLP:conf/icml/JiaYXCPPLSLD21}, and Florence~\cite{DBLP:journals/corr/abs-2111-11432} have demonstrated remarkable zero-shot generalization capabilities across a wide range of downstream tasks. 

\begin{figure}
\centering
\subfloat[]{\includegraphics[width=.31\columnwidth]{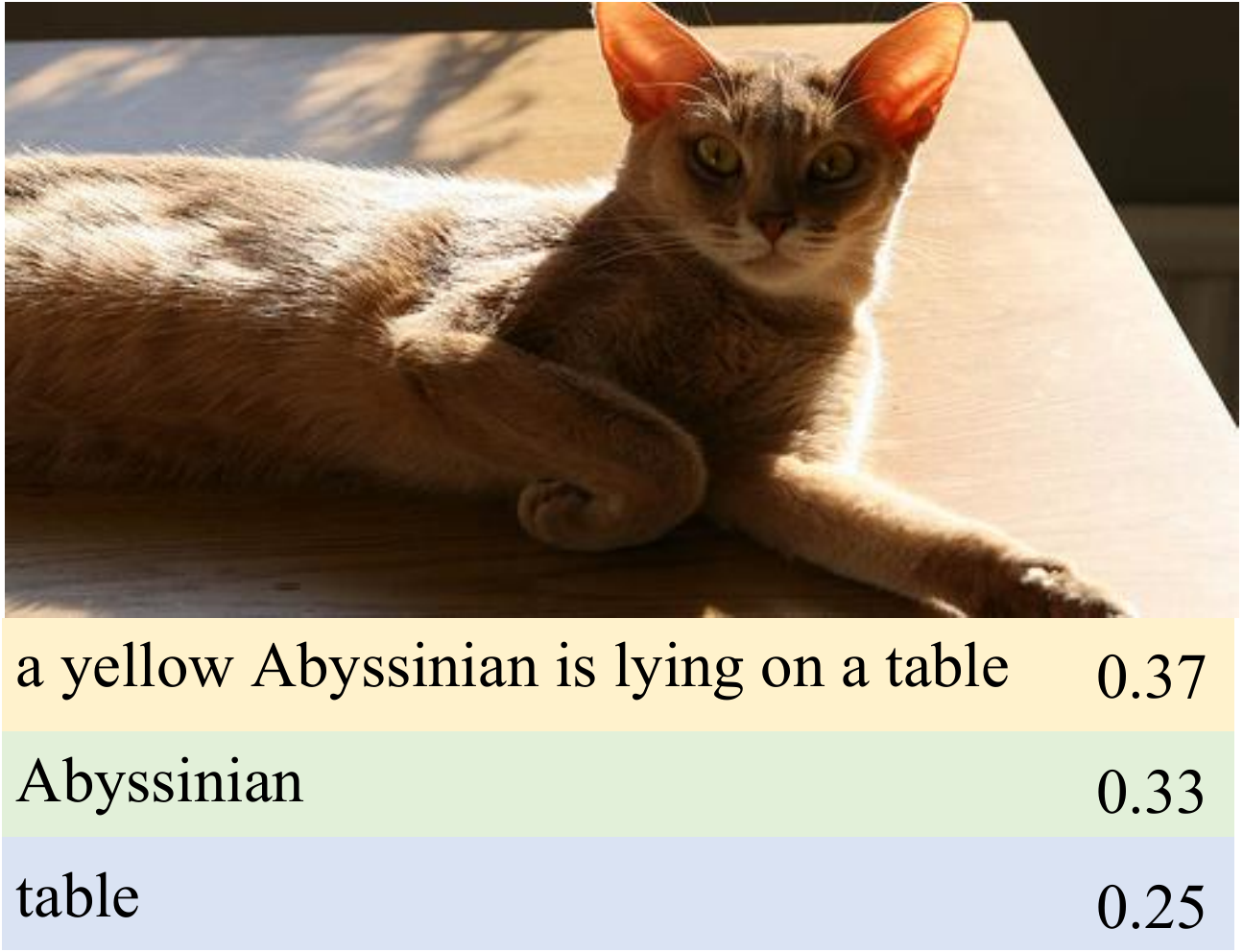}}\hspace{10.0pt}\label{fig:motivation_task}
\subfloat[]{\includegraphics[width=.402\columnwidth]{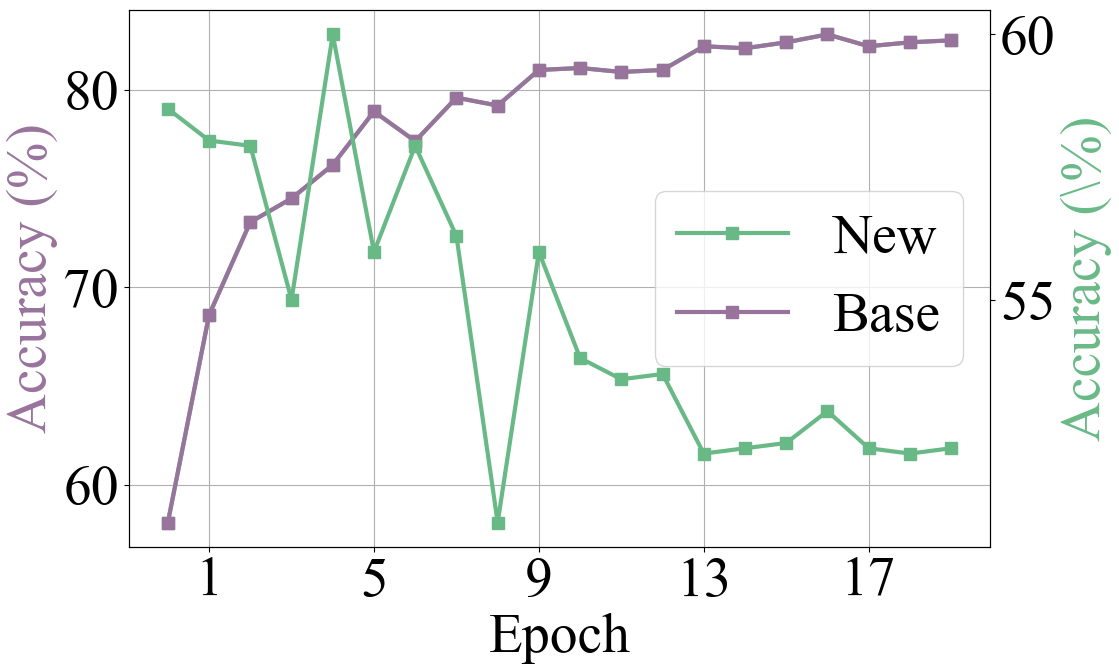}}\\
\caption{(a) A motivating example of task misalignment, illustrating the cosine similarities between an image and various text descriptions in the embedding space of CLIP. (b) A motivating experiment on data misalignment, showing the accuracy trends for base and new classes across different training epochs on the DTD dataset.}
\label{fig:motivation}
\end{figure}

Despite the strong generalization of CLIP, there exists a \textit{two-level misalignment} between CLIP and downstream tasks, which hinders its adaptation to these tasks. Specifically, the first \textit{misalignment} is caused by the discrepancy between the pre-training objectives of CLIP and the objectives of downstream tasks, i.e., the \textit{task misalignment}. During pre-training, the texts associated with images often contain rich semantic information, e.g., ``a yellow Abyssinian is lying on a table''. Accordingly, as shown in Figure~\ref{fig:motivation}(a), in the learned embedding space, the distance between the image and the entire textual description can be close, indicating a strong semantic relationship, but the distance between the image and the individual semantic elements such as ``Abyssinian'' or ``table'' is not sufficiently close. When performing a downstream task that requires classifying an image as an ``Abyssinian'', the model encounters a challenge in capturing the specific semantics associated with ``Abyssinian'' and may fail to complete the task. To address this issue, soft prompt tuning~\cite{DBLP:conf/eccv/ZhangZFGLDQL22, DBLP:conf/cvpr/ZhouYL022, DBLP:conf/cvpr/KhattakR0KK23} is proposed. By introducing learnable tokens into the model and tuning them with task-specific data, CLIP can identify task-related semantics and adapt to the corresponding task. 

However, not all downstream tasks have available annotated data for prompt tuning. As a result, inconsistency exists between the training and testing data, constituting the second \textit{misalignment}, i.e., \textit{data misalignment}. The inconsistency in data misalignment can arise due to two main reasons: (1) \textit{Label Inconsistency}: The training and testing classes do not completely overlap. For instance, some classes present in the training data might not appear in the testing data and vice versa. We refer to the classes that appear in both training and testing as base classes and the classes that appear only in testing as new classes. (2) \textit{Distribution Inconsistency}: Even if they share the same class names, the distributions of the classes in the training and testing data may differ, resulting in distribution inconsistency. In such cases, the testing classes are essentially also new classes.
Researchers generally agree that prompt tuning potentially leads to the overfitting of the CLIP model to the base classes, deviating CLIP from its original behavior and thus hindering its generalization to new classes~\cite{DBLP:conf/iccv/KhattakWNK0K23, CoPrompt}.
To explore whether the overfitting holds, we conduct prompt tuning based on MaPLe~\cite{DBLP:conf/cvpr/KhattakR0KK23} on the DTD dataset~\cite{DBLP:conf/cvpr/CimpoiMKMV14}. We record the performance of the base classes and the new classes under different training epochs. As shown in Figure~\ref{fig:motivation}(b), as the number of training epochs increases, the performance on base classes gradually improves; however, the performance on the new classes first increases and then decreases. The phenomenon is consistent with the definition of overfitting. 

To explore why overfitting occurs and how it impacts the recognition of new classes, we revisit the pre-training and adaptation processes of CLIP and develop a Structural Causal Model (SCM)~\cite{pearl2009causal,glymour2016causal} to analyze the causal relationships among variables abstracted from these two processes. From the perspective of data generation, generative factors refer to the elements that control the content of images and texts. We argue that the knowledge embedded in CLIP determines the content that can be obtained from CLIP. Therefore, the knowledge in CLIP is equivalent to the generative factors it contains. Via soft prompt tuning, we expect the learned prompts to capture the semantics determined by the task-relevant generative factors and eliminate the influence of task-irrelevant generative factors. However, due to the data misalignment, the task-relevant generative factors estimated on the base classes potentially being task-irrelevant for the new classes. When estimating the causal relationship between images and the label space of new classes, task-irrelevant generative factors could interfere. In addition, task-irrelevant generative factors cannot be observed in practice, so we cannot remove their effects by adjusting for them. To solve this problem, we introduce task-related semantics as intermediate variables from images to the label space of new classes and apply the front-door adjustment to estimate the true causal relationship between the two.

To implement the front-door adjustment, we propose \methodfull (\method), consisting of \defull (\deshort) and \mefull (\meshort). Specifically, we incorporate multiple prompt templates within the model and encourage distinct templates to represent different semantics through \deshort. In addition, decoupled semantics exhibit varying classification uncertainties. Therefore, we propose \meshort, which performs the classification task independently based on each decoupled semantic and estimates the corresponding uncertainties simultaneously. Subsequently, we fuse the results to obtain more accurate and reliable predictions. Through experiments conducted in various settings, our proposed \method has effectively enhanced the performance of CLIP. 

Our \textbf{contributions} include: (i) We identify the \textit{two-level misalignment} that exists in the adaptation of CLIP to downstream tasks and illustrate that, due to the data misalignment, the prompt learned through soft prompt tuning becomes overfitted to the base classes; (ii) We develop an SCM to investigate the pre-training and adaptation processes of CLIP, revealing how the overfitting occurs and how it impacts the recognition of new classes. We discover that the task-irrelevant generative factors serve as the confounder when estimating the true causal relationship between the images and the label space of new classes; (iii) To mitigate the impact of task-irrelevant generative factors on downstream tasks, we propose \method, which implements the front-door adjustment. Through experiments on multiple datasets and various tasks, the effectiveness of \method has been demonstrated.
\section{Related Work}
\textbf{Adaptation in Vision Language models.}
To enhance the performance of vision-language models such as CLIP on downstream tasks, researchers propose to utilize few-shot adaptation. Generally, two main technical approaches have emerged: 1) adapter-based methods~\cite{DBLP:conf/eccv/ZhangZFGLDQL22,DBLP:journals/ijcv/GaoGZMFZLQ24,DBLP:journals/tmm/PengYXWX24}, which incorporate adapters into CLIP to fine-tune the features generated by CLIP, enabling better adaptation to specific tasks; 2) prompt-based methods, which primarily inject task information by appending learnable tokens to CLIP. While methods like CoOp~\cite{DBLP:conf/eccv/ZhangZFGLDQL22}, CoCoOp~\cite{DBLP:conf/cvpr/ZhouYL022}, and ProGrad~\cite{DBLP:conf/iccv/ZhuNHWZ23} add tokens to the text input layer, VPT~\cite{DBLP:conf/eccv/JiaTCCBHL22} and CAVPT~\cite{DBLP:journals/corr/abs-2208-08340} introduce tokens on the image branch as well. MaPLe~\cite{DBLP:conf/cvpr/KhattakR0KK23} and UPT~\cite{DBLP:conf/emnlp/ChowdhuryNM23} incorporate tokens on both the image encoder and text encoder. In addition, MaPLe adds tokens into the intermediate layers of the network to enhance the model's adaptability to the task. Among all prompt-based approaches, the most similar to our method is ArGue~\cite{DBLP:journals/corr/abs-2311-16494}, which aims to improve model performance by iterating over the attributes that each category possesses. However, ArGue relies on the assistance of other large language models for attribute generation and requires additional preprocessing steps. In contrast, \method achieves competitive performance with ArGue without relying on any external knowledge, solely through the end-to-end training.

\textbf{Causal Inference.} In recent years, causal inference~\cite{pearl2009causal,glymour2016causal} has been extensively employed in computer vision and multi-modal learning. Generally, researchers explore the causal inference in two promising directions: counterfactual~\cite{DBLP:conf/cvpr/ChangAG21, DBLP:conf/cvpr/YueWS0Z21, DBLP:conf/cvpr/0016YXZPZ20,DBLP:journals/pami/ChenZNZX23} and intervention~\cite{DBLP:conf/icml/JiangCKYYW0W22, DBLP:conf/icml/QiangLZ0X22, DBLP:journals/corr/abs-2208-12681, DBLP:conf/nips/YueZS020,DBLP:conf/iclr/WangSL00W23}. ~\cite{DBLP:conf/iclr/Sauer021} and ~\cite{DBLP:conf/cvpr/0016YXZPZ20, DBLP:journals/pami/ChenZNZX23} propose to generate the counterfactual images to improve the robustness of the model. ~\cite{DBLP:conf/iclr/WangSL00W23} proposes intervening in the distribution of training data to eliminate the interference of the spurious correlation on predictions for domain generalization problems. To the best of our knowledge, we are the first to apply causal inference to analyze CLIP's overfitting in downstream classification. We identify the confounders and eliminate their effects through the front-door adjustment.
\section{Problem Formulation and Analysis}
\subsection{Problem Formulation}
CLIP~\cite{DBLP:conf/icml/RadfordKHRGASAM21} is an impressive vision-language model comprised of an image encoder and a text encoder, capable of embedding images and texts into a shared multi-modal latent space. Through contrastive learning, paired image-text embeddings are optimized for maximum cosine similarity, while unpaired ones are minimized. The intuition behind such design is based on a fundamental assumption: any paired images and texts are expected to form a tight semantic correspondence, as they contain similar content. By pre-training on a large-scale dataset consisting of 400 million image-text pairs, CLIP~\cite{DBLP:conf/icml/RadfordKHRGASAM21} demonstrates strong generalization in zero-shot and few-shot downstream tasks.

In zero-shot classification with $C$ classes, an image $x$ is transformed into an embedding $v$ via the image encoder. For the text branch, CLIP~\cite{DBLP:conf/icml/RadfordKHRGASAM21} constructs prompts for each class using a static template, e.g., ``a photo of a [CLASS]'', where [CLASS] denotes the class name of the $c$-$th$ class, $c\in \{1, 2, ..., C\}$. These prompts are inputted into the text encoder to obtain embedding vectors $w_c$ for each class. By calculating the cosine similarity between the image embedding $v$ and the text embeddings $w_c$, we can obtain the probability of image $x$ belonging to the $c$-$th$ class:
\begin{equation}
p(y=c|x)=\frac{exp(sim(v,w_c)/\tau)}{\Sigma_{c'=1}^C exp(sim(v,w_{c'})/\tau)},
\end{equation}
where $sim(\cdot, \cdot)$ denotes the cosine similarity, and $\tau$ is a temperature scalar.

In the few-shot setting, where a certain amount of labeled data is available, it is feasible to learn a template instead of manually specifying one. Initially, researchers construct a learnable template $t = \{p_1, p_2, ..., p_d, l_c\}$ by appending the word embedding $l_c$ of the $c$-$th$ class to $d$ learnable tokens as the input of the text encoder. In recent years, intermediate layers of both text and image encoders also employ learnable tokens. These tokens are optimized with the cross-entropy loss to adapt to the specific task. The learned template performs the classification task in the same way as the static one.

\subsection{Problem Analysis: Causal Perspective}\label{sec:causal}
\begin{wrapfigure}{r}{0pt}
  \centering
  \includegraphics[width=0.4\linewidth]{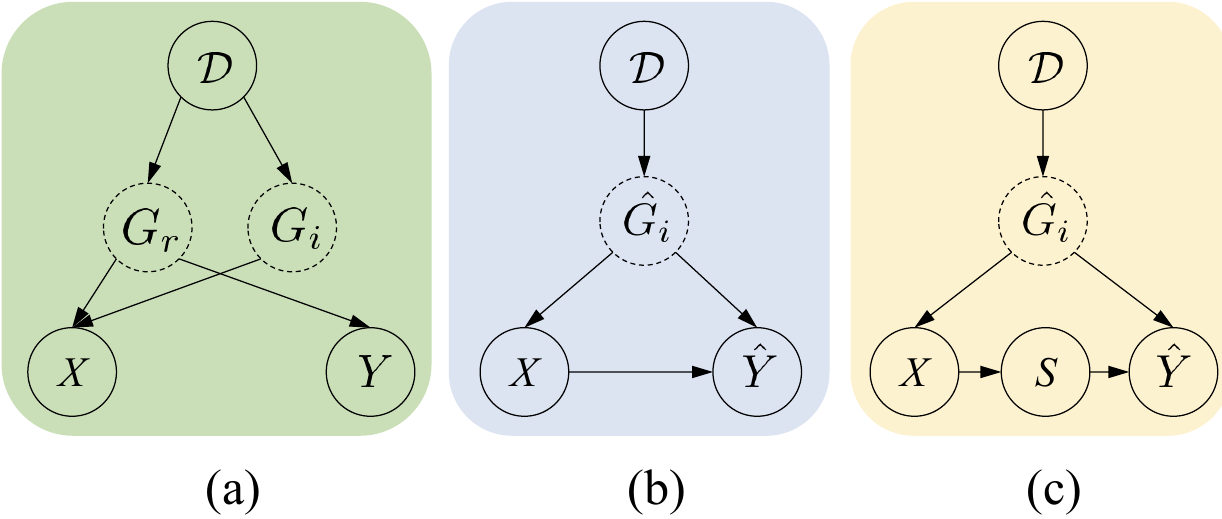}
  \caption{SCMs. Solid and dashed circles indicate the observable and unobservable variables, respectively.}
  \label{fig:scm}
\end{wrapfigure}
To understand how the knowledge contained in CLIP~\cite{DBLP:conf/icml/RadfordKHRGASAM21} affects the downstream classification tasks, we propose an SCM shown in Figure \ref{fig:scm}. The directed acyclic graph $G=<V, E>$ depicts the causal relationships among variables abstracted from the pre-training and adaptation processes. Each node $V_i\in V$ corresponds to a variable. Edge $V_i\to V_j\in E$ signifies that $V_i$ is the cause of $V_j$.

Firstly, we define each variable in Figure \ref{fig:scm}. $\mathcal{D}$ represents the data used in the pre-training stage. Generative factors, the fundamental semantic elements that control the generation of images and texts, can determine the content of both images and texts. In a specific task, these generative factors can be divided into two subsets: the set of task-relevant factors $G_r$ and the set of task-irrelevant factors $G_i$. Taking the pet classification task as an example, task-relevant generative factors determine features such as coat colors, body shapes, and ear types, which are highly relevant to identifying pet breeds. Conversely, task-irrelevant generative factors influence semantics which do not provide helpful information for pet classification. For a downstream task, $X$ represents the image space, while $Y$ corresponds to the true labels of these images. $\hat{Y}$ represents the predicted labels obtained through the learning. Additionally, $S$ denotes the semantics within the images $X$ closely relevant to the current task.

By training on diverse data, CLIP acquires relatively comprehensive visual and language knowledge. Therefore, it can be assumed that through the learning of $\mathcal{D}$, we obtain comprehensive generative factors that constitute the entire visual space. Since the learning process of pre-training is not specific to any particular downstream task, these generative factors include not only task-related but also task-irrelevant ones, i.e., $\mathcal{D}\to G_i$ and $\mathcal{D}\to G_r$, for a specific task. By combining the generative factors, any image $x\in X$ can be generated, regardless of whether it has appeared in the pre-training dataset $\mathcal{D}$. Images typically include not only content related to their category under a particular task but usually other content as well, hence $G_r\to X$ and $G_i\to X$. However, only task-related generative factors can determine the true label $Y$ of images, i.e., $G_r\to Y$.

Given that the training data $\mathcal{D}$ for CLIP is invariant, in Figure \ref{fig:scm}(a), the path $X\gets G_i\gets \mathcal{D} \to G_r \to Y$ is blocked, resulting in $X$ being associated with $Y$ solely through $G_r$, i.e., $X\gets G_r \to Y$. When performing downstream classification tasks, we aim to model the relationship between $X$ and $\hat{Y}$ that arises from $G_r$. As shown in Figure~\ref{fig:scm}(b), $X\to \hat{Y}$ represents the objective of our modeling. $G_i$ and $G_r$ are unobservable and easily confused in practice. Therefore, we perform soft prompt tuning to help capture the information of $G_r$ and eliminate the influence of $G_i$. However, the estimated task-related factors are not always accurate. Furthermore, due to the data misalignment, the task-relevant generative factors estimated based on the training data of base classes are not necessarily the task-relevant factors for new classes. $\hat{G_i}$ denotes the set of task-irrelevant generative factors that are incorrectly retained, and the set of factors that are task-relevant for the base classes but task-irrelevant for the new classes. Since we incorrectly assume that $\hat{G_i}$ is task-relevant, $\hat{G_i}$ influences $\hat{Y}$, i.e., $\hat{G_i}\to \hat{Y}$. When the learned prompt template is overfitted to the base classes, generative factors in $\hat{G}$ will further increase, hindering the estimation of the true causal relationship between $X$ and $\hat{Y}$.

According to the definition of the backdoor path, when estimating the true causal relationship between $X$ and $\hat{Y}$, there exists a backdoor path $X\gets \hat{G_i}\to \hat{Y}$, where $\hat{G_i}$ serves as a confounder. Given that $\hat{G_i}$ is unobservable,  it becomes impracticable to eliminate spurious relationships caused by $\hat{G_i}$ via the back-door adjustment. To address this issue, we introduce an intermediate variable $S$ that represents task-relevant semantics and is derived from $X$, thus establishing the path $X\to S$, as shown in Figure~\ref{fig:scm}(c). The classification for $X$ can be regarded as predicting based on the semantics $S$ that $X$ possesses, thus $S\to \hat{Y}$. $S$ satisfies the front-door criterion for $(X, \hat{Y})$, and the causal effect from $X$ to $\hat{Y}$ can be calculated using the front-door formula:
\begin{equation}
P({\hat{Y}=y}|do({x})) =\sum\limits_{s} {\underbrace{\vphantom{\sum\limits_{x'}}P(s|{x})}_\text{first\  term}\underbrace{\sum\limits_{{{x'}}} {P(y|{{x'}},s)P({{x'}})}}_\text{second \ term} }.
\label{eq:frontdoor}
\end{equation}
To estimate the true causal relationship between $X$ and $\hat{Y}$, we will discuss how to obtain the semantic $s$ and how to estimate the probability of a sample $x$ belonging to class $y$. 
\begin{figure}
  \centering
  \includegraphics[width=0.85\linewidth]{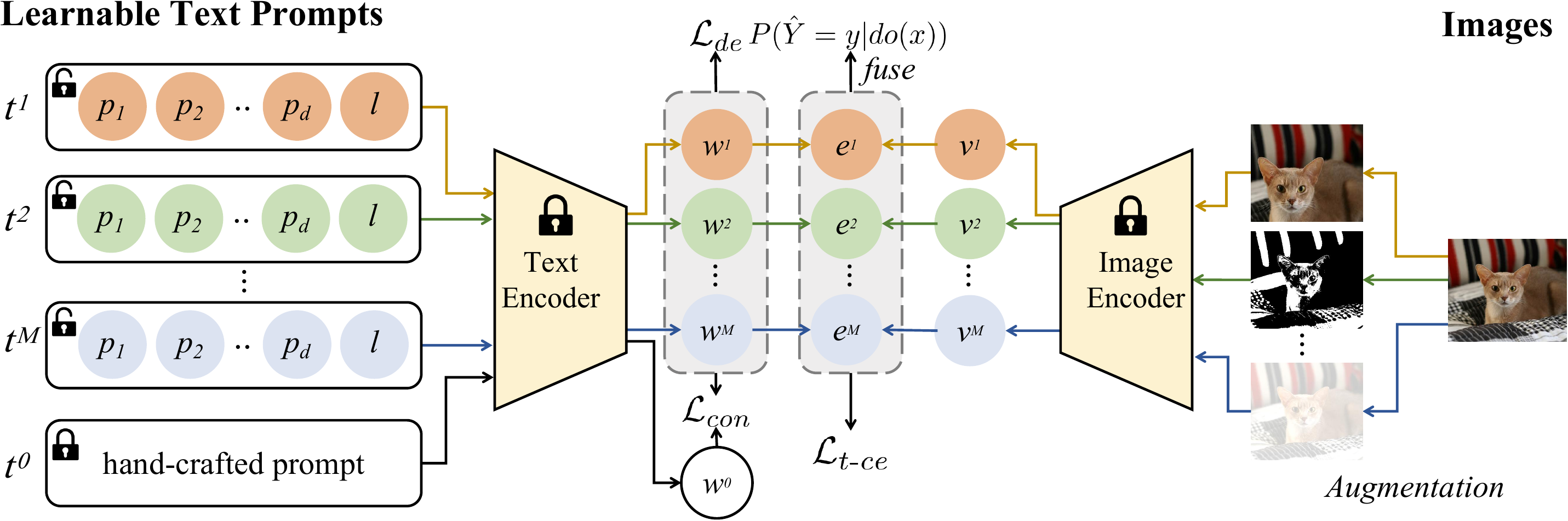}
  \caption{Framework of \method. $t^m$ denotes a single template, while $p_1, p_2, ..., p_d$ represent tokens in the template. Different colors indicate diverse templates. ``fuse'' refers to the process of generating the final classification results from multiple template results as shown in Equation \eqref{eq:cdc_totalfuse}. The text encoder and the image encoder are frozen, and only the tokens in the prompt templates are learnable. 
  }
  \label{fig:framework}
\end{figure}
\section{Methodology}\label{sec:method}
Guided by the front-door criterion, we propose \method, comprising two essential components: \deshort and \meshort. Specifically, \deshort focuses on decoupling the semantics represented by diverse templates. \meshort aims to generate the classification scores via Equation~\eqref{eq:frontdoor} based on the decoupled semantics. We will elaborate on the details of these two components in Section \ref{sec:de} and Section \ref{sec:me}, respectively.
\subsection{\defull}\label{sec:de}
To decouple the semantics embedded in the images and texts, the concept of \textit{semantic set} is introduced. A semantic set serves as a collection that encapsulates the semantics related to the same visual attribute. For example, multiple semantic sets can be constructed for the task of pet classification, such as the set of coat colors, the set of body shapes, and the set of ear types. Each semantic set contains some possible values that represent the variations of the corresponding attribute. For instance, the semantic set ``coat colors'' includes semantics like red, white, purple, etc. The semantic sets enable us to understand and analyze the semantic information embedded in pet images.

To represent different semantic sets, we employ multiple templates in the CLIP model, i.e., $t^m\in \{t^1, t^2, ..., t^M\}$ in Figure~\ref{fig:framework}, where $M$ denotes the number of templates. Each template aims to characterize a unique semantic set. 
To achieve decoupling of the semantics represented by these templates, we propose \deshort, which considers both text and image branches. On the image branch, we leverage diverse augmentation methods to generate the image inputs and employ the corresponding augmentation method for each individual template. The intuition behind this is to help different templates capture distinct feature invariance against different augmentations, e.g., performing random rotation augmentation enables the template to capture features with rotation invariance. Since defining the precious semantic sets, such as coat colors and ear types, remains challenging without the help of external knowledge, we group semantics that exhibit the same type of feature invariance into a semantic set in practice.

On the text branch, we expect to maximize the diversity of embeddings corresponding to different templates, ideally achieving orthogonality. To achieve this objective, we aim to ensure that the embedding of the $c$-$th$ class in the $m$-$th$ template, denoted as $w^m_c$, cannot be accurately classified by embeddings from other templates, i.e., using any $w^{m'}$ to classify $w^m_c$, where $m'\neq m$, there is no bias towards the results. Specifically, the objective function can be formalized as:
\begin{equation}
\mathcal{L}_{de}=\frac{1}{M-1}\frac{1}{C}\sum_{m'=1,m'\neq m}^{M}\sum_{c=1}^{C}\sum_{\bar{c}=1}^{C}P(\bar{c}|w^m_c, w^{m'})\log P(\bar{c}|w^m_c, w^{m'}),
\label{eq:entropy}
\end{equation}
where $P(\bar{c}|w^m_c, w^{m'})$ denotes the probability that $w^m_c$ is predicted as the $\bar{c}$-$th$ class by the embedding of $m'$-$th$ template. Specifically, $P(\bar{c}|w^m_c, w^{m'})$ can be calculated as:
\begin{equation}
P(\bar{c}|w^m_c, w^{m'})=\frac{exp(sim(w_c^m,w^{m'}_{\bar{c}})/\tau)}{\sum_{c'=1}^C exp(sim(w_c^m,w^{m'}_{c'})/\tau)}.
\label{eq:p_calculate}
\end{equation}
Intuitively, the embedding vector $w_c^m$ from the $m$-$th$ template can be regarded as a sample to be classified. The objective of Equation \eqref{eq:entropy} is to maximize the entropy of the classification results, resulting in a uniform distribution. The classification results are obtained using the embedding of all other templates. $\mathcal{L}_{de}$ aims to enhance the semantic distinctions among different templates, thereby encouraging each template to maintain a unique semantic.

While boosting the diversity of the semantics in different templates, $\mathcal{L}_{de}$ may hinder the learning of the task-relevant semantics. To alleviate this problem, we propose introducing a consistency loss to assist each template in capturing information that is close to the hand-crafted template. The proposed consistency loss can be formulated as follows:
\begin{equation}
\mathcal{L}_{con}=-\frac{1}{C}\sum_{m=1}^{M}\sum_{c=1}^{C}\log P(c|w_c^m,w^0),
\label{eq:con}
\end{equation}
where $w^0$ denotes the embeddings of the hand-crafted template $t^0$ obtained via the original CLIP model. See \textbf{Appendix~\ref{app:prompts}} for the detailed templates.

Via \deshort, we obtain distinct semantic sets that contain decoupled semantics. Different classes have varying values on the same semantic set, so we consider the values of $s$ in Equation \eqref{eq:frontdoor} to be the text embeddings of each class on all semantic sets, i.e, $s\in \{w^m_c\}_{m=1,c=1}^{M, C}$. 

\subsection{\mefull}\label{sec:me}
To implement Equation~\eqref{eq:frontdoor}, we need to clarify how to estimate the $P(\hat{Y}|do(x))$ based on the decoupled semantics. Firstly, the probability $P(s|x)$ that a sample $x$ has semantic $s$ is obtained by calculating the cosine similarity between the representation $v$ of $x$ and $s$, i.e., $P(s|x)=N(sim(s,v))$, where $N$ denotes the normalization operation.

After determining the possible values of $s$, we clarify how to estimate the second term in Equation \eqref{eq:frontdoor}. Let $x'$ represent the samples from the training set and $s = w_c^m$, then: (i) if $c\neq y$, $P(y|x',s)=0$. By updating the template, we expect each class $c$ to carry unique semantics $w_c^m$. Therefore, it cannot be assumed that the specific semantics of $c$ is attributable to another class; (ii) if $c=y$ but $x'$ does not belong to the $c$-$th$ class, $P(y|x',s)=0$; (iii) if $c=y$ and $x'$ belong to the $c$-$th$ class, we maximize the similarity between $s$ and $x'$ by prompt tuning so that the semantics of $s$ and $x'$ are as consistent as possible. As $x'$ contains both task-relevant and task-irrelevant semantics, and its task-relevant semantic is decoupled to $w_c^1, w_c^2, ..., w_c^M$ via \deshort, $s$ is the subset of $x'$. Therefore, the second term in Equation~\eqref{eq:frontdoor} shrinks to $P(y|s)$.

Ideally, if $s$ contains and only contains the category-specific semantic information that explicitly supports the classification of sample $x'$ into $c$,  $P(y=c|s)=1$, and $P(y\neq c|x',s)=0$. However, because of the data misalignment, while applying the learned templates to the new classes, this ideal scenario does not happen. In reality, the qualities of the learned semantics vary for different templates and classes, allowing the value of $P(y|s)$ to alter, thereby resulting in significant differences in the uncertainty of classification results derived from different templates. To solve this issue, we propose to estimate the uncertainty~\cite{DBLP:conf/iclr/HanZFZ21,DBLP:journals/pami/HanZFZ23} of each prediction from diverse templates via Dempster-Shafer theory~\cite{shafer1976mathematical, dempster2008upper}. Specifically, we consider $e_c^m=h(sim(w_c^m,v))$ as the evidence of classifying the feature $v$ as belonging to the $c$-$th$ class under the $m$-$th$ semantic set, where $h$ represents a function that maps the cosine similarity to a non-negative value. We assume that there exists a Dirichlet distribution $\mathbf{\alpha}^m=\left[\alpha_1^m, \alpha_2^m,..., \alpha_C^m\right]$, where $\alpha_c^m=e_c^m+1$. This formulation allows us to quantify the belief that the feature $v$ belongs to class $c$ as well as the uncertainty associated with the prediction as:
\begin{equation}
\label{eq:belif}
b_c^m=\frac{e_c^m}{A^m}, u^m=\frac{C}{A^m},
\end{equation}
where $A^m=\sum\limits_{c'=1}^C e_{c'}^m + 1$. 
Once we have obtained the classification beliefs and uncertainties for all semantic sets, we proceed to perform evidence fusion to obtain the final classification results. Specifically, for the $m$-$th$ and $m'$-$th$ semantic set, after evidence fusion, we obtain the belief of the sample $x$ belonging to class $c$, as well as the uncertainty of the classification as:
\begin{equation}
\label{eq:fuse}
b_c^{mm'}=\frac{1}{1-C}(b_c^mb_c^{m'}+b_c^mu^{m'}+b_c^{m'}u^m),u^{mm'}=\frac{1}{1-C}u^mu^{m'}.
\end{equation}
The results from all prompt templates can be iteratively fused, as shown in the following equation:
\begin{equation}
\begin{aligned}
B_c^m &= \begin{cases}
b_c^1, & \text{if } m = 1 \\
\frac{1}{1-C}(B_c^{m-1}b_c^m + B_c^{m-1}u^m + b_c^mU^{m-1}), & \text{if } 1 < m \leq M
\end{cases}, \\
U^m &= \begin{cases}
u^1, & \text{if } m = 1 \\
\frac{1}{1-C}U^{m-1}u^m, & \text{if } 1 < m \leq M
\end{cases}.
\end{aligned}
\label{eq:cdc_totalfuse}
\end{equation} 
In this formulation, $B_c^m$ represents the belief after fusing the results from the first $m$ templates, while $U^m$ denotes the corresponding uncertainty. The process of evidence fusion can be understood as the summation of results over all semantics $s$ in Equation \eqref{eq:frontdoor}. Consequently, the result in Equation \eqref{eq:frontdoor} can be reformulated as:
\begin{equation}
P({\hat{Y}=c}|do({x})) = \frac{B_c^M}{\Sigma_{c'=1}^CB_{c'}^M}.
\label{eq:cdc_prob}
\end{equation}

To calculate the probability that sample $x$ belongs to class $c$ during the testing phase, we employ Equation \eqref{eq:cdc_prob}. Accordingly, during the training stage, the cross-entropy loss which is used for learning the prompts is adjusted to the trusted cross-entropy loss to model the uncertainty information. For each training sample, the trusted cross-entropy can be formalized as:
\begin{equation}
\mathcal{L}_{t\text{-}ce}=\sum_{m=1}^M(\psi(A)-\psi(\alpha_y^m)),
\label{eq:tce}
\end{equation}
where $y$ denotes the index of the ground-truth class and $\psi(\cdot)$ denotes the digamma function.
\subsection{Overall Architecture}
We demonstrate the architecture of our proposed method in Figure~\ref{fig:framework}. While training, the overall loss function can be formalized as:
\begin{equation}
\mathcal{L}_{\method}=\mathcal{L}_{t\text{-}ce}+\beta \mathcal{L}_{de}+\gamma \mathcal{L}_{con},
\label{eq:loss}
\end{equation}
where $\beta$ and $\gamma$ denote the hyper-parameters to tune the influence of $\mathcal{L}_{de}$ and $\mathcal{L}_{con}$, respectively. Refer to \textbf{Algorithm \ref{alg:training}} and \textbf{Algorithm \ref{alg:testing}} for the detailed training and testing pipeline.

\begin{algorithm}
\begin{algorithmic}
\STATE {\bfseries Input:} 16-shot dataset $X$, the learning rate $\ell$, two hyper-parameter $\beta$ and $\gamma$, the number of templates $M$, and the list of augmentation methods $\{\mathcal{A}^1, \mathcal{A}^2, ..., \mathcal{A}^M\}$.\\
\STATE Initialize the parameters of CLIP with the parameters of the pre-trained model.
\STATE Randomly initialize the learnable tokens $\{\theta^1, \theta^2, ..., \theta^M\}$.
\REPEAT
\FOR{$i$-$th$ training iteration}
\STATE Iteratively sample a minibatch $\mathcal{X}^\prime$ from $\mathcal{X}$.
\FOR{$m$-$th$ template}
\STATE ${\mathcal{X}^\prime}^m=\mathcal{A}^m(\mathcal{X}^\prime)$.
\STATE Generate the image features $v^m$ of ${\mathcal{X}^\prime}^m$ with the image encoder of CLIP.
\STATE Generate the text features $w^m$ with the text encoder.
\STATE Generate the classification evidence $e^m$ with $v^m$ and $w^m$: $e_c^m=h(sim(w_c^m,v^m))$.
\STATE Using Equation \eqref{eq:tce} to calculate $\mathcal{L}_{t\text{-}ce}$.
\STATE Using Equation \eqref{eq:entropy} and Equation \eqref{eq:con} to calculate $\mathcal{L}_{de}$ and $\mathcal{L}_{con}$.
\STATE Using Equation \eqref{eq:loss} to calculate $\mathcal{L}_{\method}$.
\STATE $\theta^m=\theta^m-\ell\nabla_{\theta} \mathcal{L}_{\method}$.
\ENDFOR
\ENDFOR
\UNTIL $\theta$ converge.
\end{algorithmic}
\caption{The training pipeline of \method}
\label{alg:training}
\end{algorithm}
\vspace{6.0pt}
\begin{algorithm}
\begin{algorithmic}
\STATE {\bfseries Input:} The testing dataset $X$, the number of templates $M$, and $\{\theta^1, \theta^2, ..., \theta^M\}$.
\STATE Sample a test data $x$ from $X$
\FOR{$m$-$th$ template}
\STATE Generate the image features $v^m$ of $x$.
\STATE Generate the text features $w^m$ with the text encoder.
\STATE Generate the classification evidence $e^m$ with $v^m$ and $w^m$: $e_c^m=h(sim(w_c^m,v^m))$.
\ENDFOR
\STATE Using Equation \eqref{eq:cdc_totalfuse} and Equation \eqref{eq:cdc_prob} to generate the final classification results.
\end{algorithmic}
\caption{The testing pipeline of \method}
\label{alg:testing}
\end{algorithm}

\section{Experiments}
Following previous works~\cite{DBLP:conf/cvpr/ZhouYL022,DBLP:conf/cvpr/KhattakR0KK23}, we conduct experiments to evaluate our proposed method with three different settings. These settings encompass the base-to-new setting as well as two out-of-distribution (OOD) settings, i.e., the cross-dataset setting and the cross-domain setting. Refer to \textbf{Appendix~\ref{app:settings}} for a detailed overview of the evaluation protocol.

\subsection{Experimental Settings}\label{sec:exp_details}
\textbf{Datasets.} In the base-to-new setting, we conduct experiments based on 11 datasets: ImageNet~\cite{DBLP:conf/cvpr/DengDSLL009}, Caltech101~\cite{DBLP:journals/cviu/Fei-FeiFP07}, Oxford Pets~\cite{DBLP:conf/cvpr/ParkhiVZJ12}, Stanford Cars~\cite{DBLP:conf/iccvw/Krause0DF13}, Flowers102~\cite{DBLP:conf/icvgip/NilsbackZ08}, Food101~\cite{DBLP:conf/eccv/BossardGG14}, FGVC Aircraft~\cite{DBLP:journals/corr/MajiRKBV13}, SUN397~\cite{DBLP:conf/cvpr/XiaoHEOT10}, DTD~\cite{DBLP:conf/cvpr/CimpoiMKMV14}, EuroSAT~\cite{DBLP:journals/staeors/HelberBDB19}, and UCF-101~\cite{DBLP:journals/corr/abs-1212-0402}. In the following, we abbreviate these datasets as ImageNet, Caltech, Pets, Cars, Flowers, Food, Aircraft, SUN, DTD, EuroSAT, and UCF. For the out-of-generalization task, we adopt ImageNet as the source dataset. The remaining 10 are the target datasets in the cross-dataset setting, and ImageNetV2~\cite{DBLP:conf/icml/RechtRSS19}, ImageNet-S~\cite{DBLP:conf/nips/WangGLX19}, ImageNet-A~\cite{DBLP:conf/cvpr/HendrycksZBSS21}, and ImageNet-R~\cite{DBLP:conf/iccv/HendrycksBMKWDD21} are the target datasets in the cross-domain setting.

\begin{table}[t]
\caption{The comparison with baseline methods on base-to-novel generalization setting.}
\centering
\setlength{\tabcolsep}{0.8mm}
\begin{tabular}{l|ccc|ccc|ccc|cccc}
\toprule
\multirow{2}{*}{Dataset} & \multicolumn{3}{c|}{CoOp~\cite{DBLP:conf/eccv/ZhangZFGLDQL22}}      & \multicolumn{3}{c|}{CoCoOp~\cite{DBLP:conf/cvpr/ZhouYL022}} & \multicolumn{3}{c|}{MaPLe~\cite{DBLP:conf/cvpr/KhattakR0KK23}} & \multicolumn{4}{c}{\method} \\ \cmidrule{2-14} 
& Base & New & HM & Base & New & HM & Base & New & HM & Base & New & HM & $\Delta$ \\ 
\midrule 
Avg & 82.69 & 63.22 & 71.66 & 80.47 & 71.69 & 75.83 & 82.28 & 75.14 & 78.55 & \textbf{83.34} & \textbf{77.38} & \textbf{80.25} & +1.70   \\
\midrule
ImageNet & 76.47 & 67.88 & 71.92 & 75.98 & 70.43 & 73.10 & 76.66 & 70.54 & 73.47 & \textbf{77.50} & \textbf{71.73} & \textbf{74.51} & +1.04\\ 
Caltech & 98.00 & 89.91 & 93.73 & 97.96 & 93.81 & 95.84 & 97.74 & 94.36 & 96.02 & \textbf{98.20} & \textbf{94.37} & \textbf{96.25} & +0.23\\ 
Pets & 93.67 & 95.29 & 94.47 & 95.20 & 97.69 & 96.43 & 95.43 & 97.76 & 96.58 & \textbf{96.07} & \textbf{98.00} & \textbf{97.02} & +0.44\\ 
Cars & \textbf{78.12} & 60.40 & 68.13 & 70.49 & 73.59 & 72.01 & 72.94 & \textbf{74.00} & 73.47 & 73.80 & 73.97 & \textbf{73.88} & +0.41 \\ 
Flowers & \textbf{97.60} & 59.67 & 74.06 & 94.87 & 71.75 & 81.71 & 95.92 & 72.46 & 82.56 & 96.93 & \textbf{75.07} & \textbf{84.61} & +2.05 \\ 
Food & 88.33 & 82.26 & 85.19 & 90.70 & 91.29 & 90.99 & 90.71 & 92.05 & 91.38 & \textbf{90.87} & \textbf{92.33} & \textbf{91.59} & +0.21 \\ 
Aircraft & \textbf{40.44} & 22.30 & 28.75 & 33.41 & 23.71 & 27.74 & 37.44 & 35.61 & 36.50 & 37.47 & \textbf{37.50} & \textbf{37.48} & +0.98 \\ 
SUN & 80.60 & 65.89 & 72.51 & 79.74 & 76.86 & 78.27 & 80.82 & 78.70 & 79.75 & \textbf{82.37} & \textbf{80.03} & \textbf{81.18} & +1.43 \\ 
DTD & 79.44 & 41.18 & 54.24 & 77.01 & 56.00 & 64.85 & 80.36 & 59.18 & 68.16 & \textbf{82.70} & \textbf{64.10} & \textbf{72.22} & +4.06 \\ 
SAT & 92.19 & 54.74 & 68.90 & 87.49 & 60.04 & 71.21 & 94.07 & 73.23 & 82.35 & \textbf{95.10} & \textbf{82.33} & \textbf{88.26} & +5.91 \\ 
UCF & 84.69 & 56.05 & 67.46 & 82.33 & 73.45 & 77.64 & 83.00 & 78.66 & 80.77 & \textbf{85.70} & \textbf{81.73} & \textbf{83.67} & +2.90 \\ 
\bottomrule
\end{tabular}
\label{tab:base2new}
\end{table}

\begin{table}
\caption{ Comparison of \method with recent approaches on cross-dataset evaluation. }
\setlength{\tabcolsep}{0.5mm}
\scalebox{1.00}{
\begin{tabular}{l c ccccccccccc}
\toprule
& \textbf{Source} & \multicolumn{11}{c}{\textbf{Target}} \\ \cmidrule(lr){2-2} \cmidrule(lr){3-13}
& {ImageNet} & {Caltech} & {Pets} & {Cars} & {Flowers} & {Food} & {Aircraft} & {SUN} & {DTD} & {SAT} & {UCF} &{{Avg}} \\
\midrule
CoOp & {71.51} & 93.70 & 89.14 & 64.51 & 68.71 & 85.30 & 18.47 & 64.15 & 41.92 & {46.39} & 66.55 & 63.88 \\
Co-CoOp & 71.02 & {94.43} & 90.14 & 65.32 & 71.88 & 86.06 & 22.94 & {67.36} & 45.73 & 45.37 & 68.21 & 65.74 \\
MaPLe & 70.72 & 93.53 & {90.49} & 65.57 & {72.23} & {86.20} & \textbf{24.74} & 67.01 & {46.49} & {48.06} & \textbf{68.69} & 66.30 \\
\midrule
\method & \textbf{71.76} & \textbf{94.47} & \textbf{90.77} & \textbf{66.27} & \textbf{72.67} & \textbf{86.27} & 24.50 & \textbf{68.07} & \textbf{46.60} & \textbf{49.13} & 68.60 & \textbf{66.73}  \\
\bottomrule
\end{tabular}}
\label{tab:cross_dataset}
\end{table}
\textbf{Experimental Details.} We follow the experimental settings of the baseline method MaPLe. Specifically, we utilize a pre-trained CLIP with ViT-B/16 as the visual encoder. The number of learnable tokens is fixed at 2, whereas the prompt depth varies, being 9 for the base-to-new setting and 3 for the OOD setting. The learning rate is 0.035, and the batch size is 4. All models are trained using an SGD optimizer on an NVIDIA 3090 GPU. Our proposed \method introduces three additional hyperparameters: $\beta$ and $\gamma$, which represent the weights for $\mathcal{L}_ {de}$ and $\mathcal{L}_{con}$, respectively, and $M$, which denotes the number of prompts. Furthermore, different augmentation methods have a notable impact on the model's performance. We set $\beta=5$, $\gamma=0.01$, and $M=4$ in the base-to-new setting, and $\beta=3$, $\gamma=0.01$, and $M=8$ in the OOD setting. For a detailed analysis of the influence of varying hyper-parameters and augmentation methods, please refer to \textbf{Appendix~\ref{app:exp}}. We report the accuracies of base classes and novel classes, along with their harmonic mean (HM), averaged over 3 runs. 
\subsection{Base-to-New Generalization}
As shown in Table~\ref{tab:base2new}, we compare our method with three recent works, CoOp, CoCoOp, and MaPLe. For a fair comparison, we do not include methods that utilize external knowledge, such as ArGue~\cite{DBLP:journals/corr/abs-2311-16494}, HPT~\cite{DBLP:conf/aaai/WangJCLZ24}, and CoPrompt~\cite{CoPrompt}, although our method demonstrates competitive performance.

Compared to the baseline method MaPLe, our approach has achieved remarkable improvements in the accuracy of base classes by 1.06\%, and the accuracy of new classes by 2.24\%,  which leads to an improvement of 1.70\% in HM. Specifically, across all 11 datasets, our method outperforms MaPLe in all 11 datasets for base classes and in 10 datasets for novel classes. Notably, on the ImageNet dataset, our approach achieves an accuracy increase of 0.84\% in base classes, 1.19\% in new classes, and 1.04\% in HM. For the challenging datasets SAT, DTD, and UCF, our method delivers significant HM improvements of 5.91\%, 4.06\%, and 2.90\%, respectively. In summary, our proposed method significantly enhances the generalization of the CLIP model for unseen classes, while ensuring stable improvements in the performance of base classes, demonstrating its superiority on a wide range of datasets.

\subsection{Out-of-Distribution Generalization}
\textbf{Cross-dataset.} We evaluate the cross-dataset generalization of \method by learning prompts on ImageNet and then transferring them to the remaining 10 datasets. From Table~\ref{tab:cross_dataset}, for the source dataset ImageNet, our proposed method achieves performance improvement by 1.04\% compared to MaPLe, outperforming all previous methods. Compared to MaPLe on the 10 target datasets, our method achieves better performance on 8 datasets. Considering all target datasets, our method achieves an average performance improvement of 0.43\%.

\textbf{Cross-domain. }We transfer the prompts learning in ImageNet to its four invariants to evaluate the coss-domain generalization of \method. From Table~\ref{tab:cross_domain}, compared to MaPLe, our proposed method improves the performance of ImageNetV2, ImageNet-S, and ImageNet-R by 0.80\%, 1.18\%, and 1.12\%, respectively, and brings a slight performance decrease of 0.5\% for ImageNet-A, thus leading to an average performance improvement of 0.65\%. Generally, \method enhances the generalization of the model when dealing with data from different domains.

\begin{table}
\centering
\caption{ Comparison of \method with recent approaches in the cross-domain setting. }
\label{tab:cross_domain}
\begin{tabular}{l cccccc}
\toprule
& \textbf{Source} & \multicolumn{4}{c}{\textbf{Target}} \\ \cmidrule(lr){2-2} \cmidrule(lr){3-7}
& ImageNet & ImageNetV2 & ImageNet-S & ImageNet-A & ImageNet-R & Avg.\\
\midrule
CLIP &  66.73 & 60.83 & {46.15} & 47.77 & {73.96} & 57.18 \\
CoOp &  71.51 & {64.20} & 47.99  & 49.71  & 75.21 & 59.28 \\
Co-CoOp & 71.02 & {64.07} & 48.75 & 50.63 & 76.18 & 59.91 \\
MaPLe & 70.72 & {64.07} & {49.15} & \textbf{50.90}  & {76.98} & 60.28 \\
\midrule
\method & \textbf{71.76} & \textbf{64.87} & \textbf{50.33} & 50.40 & \textbf{78.10} & \textbf{60.93} \\
\bottomrule
\end{tabular}
\end{table}

\subsection{Ablative Experiments}\label{sec:ablative}
Table \ref{tab:ablation} presents the experimental results obtained after adding some of the components to the baseline method MaPLe. The performance of MaPLe is shown in the first row. 

\begin{wraptable}[14]{r}{7.0cm}
\caption{The results of the ablative experiments on base-to-novel generalization setting.}
\centering
\setlength{\tabcolsep}{0.8mm}
\begin{tabular}{ccccccc}
\toprule
\multirow{2}{*}{M} & \multirow{2}{*}{\meshort}  & \multicolumn{2}{c}{\deshort} & \multirow{2}{*}{Base} & \multirow{2}{*}{New }& \multirow{2}{*}{HM} \\ \cmidrule{3-4} 
& & Image & Text \\ 
\midrule 
1 & $\times$ & $\times$ & $\times$ & 82.28 & 75.14 & 78.55 \\
\midrule
4 & $\times$ & $\times$ & $\times$ & 82.96 & 76.06 & 79.36 \\
4 & \checkmark & $\times$ & $\times$ & 82.65 & 76.78 & 79.61 \\
4 & \checkmark & \checkmark & $\times$ & \textbf{83.35} & 76.72 & 79.90 \\
4 & \checkmark & $\times$ & \checkmark & 82.93 & 76.83 & 79.77 \\
4 & \checkmark & \checkmark & \checkmark & 83.34 & \textbf{77.38 }& \textbf{80.25} \\
\bottomrule
\end{tabular}
\label{tab:ablation}
\end{wraptable}
\textbf{The effectiveness of multiple templates and \meshort. }In Table \ref{tab:ablation}, all results except the baseline method are obtained by increasing the number of templates to four. In the second row, we train the learnable tokens within the templates using the conventional cross-entropy loss and average the predictions from different templates to generate the final classification results. Compared to the baseline, multiple templates lead to a 0.68\% improvement in the performance of base classes and a 0.92\% enhancement in the performance of new classes, resulting in an overall HM gain of 0.81\%. While modeling the uncertainty of each template, as shown in the third line, the performance for new classes achieves an improvement of 0.72\%, and HM increases by 0.25\%. Without explicitly decoupling the semantics within different templates, increasing the number of templates effectively boosts the performance. 
We argue that the semantics acquired by distinct templates, each with varied initialization, is inherently independent (See analysis in \textbf{Appendix~\ref{app:exp}}). Therefore, simply applying multiple templates guided by the front-door adjustment could enhance the performance.

\textbf{The effectiveness of \deshort. }The fourth and fifth rows of Table \ref{tab:ablation} present the experimental results obtained by decoupling semantics solely within the image and text branches, respectively. When compared to the third row, it becomes evident that decoupling semantics at either branch alone does not yield a substantial improvement in the accuracy of new classes. However, upon comparing the sixth row with the third row, it is observed that when both decoupling methods are employed concurrently, the model achieves a 0.69\% increase in base class performance and a 0.60\% improvement in novel class performance, leading to an overall HM enhancement of 0.64\%. Despite the inherent decoupling of semantics among different templates, the utilization of \deshort further promotes the learning of distinct semantics for each template, thereby enhancing the model's overall performance.
\section{Conclusion and Future Discussion} \label{sec:conclusion}
Through motivating experiments, we identify the two-level misalignment that exists when CLIP is applied to downstream tasks. We further show that soft prompt tuning may worsen the second misalignment due to overfitting to the base classes, which impairs the generalization of CLIP. To analyze this problem, we propose an SCM and discover that the task-irrelevant generative factors serve as the confounder while estimating the true causal relationship between images and label space of new classes. To address this issue, we propose to implement front-door adjustment via \method. Specifically, we introduce multiple templates to represent the decoupled semantics, then leverage \deshort to further facilitate the decoupling of semantics, and finally fuse predictions from different templates via \meshort. The results under multiple experimental settings demonstrate that our proposed approach effectively improves the generalization of CLIP when adapted to downstream tasks.

{\bf{Limitations and broader impacts}}. \method incurs more time consumption, as applying different templates in the image branch requires encoding an image multiple times during both training and testing. The analysis of two-level misalignment in CLIP and the modeling of SCM is inspiring for the community.

\section*{Acknowledgments}
The authors gratefully acknowledge the valuable feedback provided by the anonymous reviewers. This work is supported by the Fundamental Research Program, China, Grant No.JCKY2022130C020.

\bibliographystyle{unsrt}
\bibliography{neurips_2024}


\appendix
\newpage
\section{Hand-Crafted Prompts}\label{app:prompts}
To prevent the text embeddings obtained via the learned prompts from deviating from the semantics of their respective categories, we introduce a consistency loss $\mathcal{L}_{con}$ to apply constraints on the learned prompts. For each dataset, we borrow from CLIP a hand-crafted prompt $t^0$ to generate $w^0$. The prompts employed in this process are detailed below.
    
    \ \ \ \ \ \  \textit{``\textbf{Pets}'': ``a photo of a \{\}, a type of pet.''}
    
    \ \ \ \ \ \ \textit{``\textbf{Flowers}'': ``a photo of a \{\}, a type of flower.''}
    
    \ \ \ \ \ \ \textit{``\textbf{Aircraft}'': ``a photo of a \{\}, a type of aircraft.''}
    
    \ \ \ \ \ \ \textit{``\textbf{DTD}'': ``a photo of a \{\}, a type of texture.''}
    
    \ \ \ \ \ \ \textit{``\textbf{SAT}'': ``a centered satellite photo of \{\}.''}
    
    \ \ \ \ \ \ \textit{``\textbf{Cars}'': ``a photo of a \{\}.''}
    
    \ \ \ \ \ \ \textit{``\textbf{Food}'': ``a photo of \{\}, a type of food.''}
    
    \ \ \ \ \ \ \textit{``\textbf{SUN}'': ``a photo of a \{\}.''}
    
    \ \ \ \ \ \ \textit{``\textbf{Caltech}'': ``a photo of a \{\}.''}
    
    \ \ \ \ \ \ \textit{``\textbf{UCF}'': ``a photo of a person doing \{\}.''}
    
    \ \ \ \ \ \ \textit{``\textbf{ImageNet}'': ''a photo of \{\}.''}
\section{Evaluation Protocol}\label{app:settings}
\textbf{Base-to-New Setting.} Following CoCoOp and MaPLe, we split the classes into two un-overlapping subsets, i.e., base classes and new classes, for each dataset. In training, base classes are leveraged for learning the prompts under the 16-shot setting, where each class contains 16 annotated samples. Subsequently, the prompts are transferred to new classes which are unseen during the training. In this setting, the evaluation metrics include accuracy on base classes, new classes, and their corresponding harmonic mean (HM), which can be calculated by:
\begin{equation}
HM=\frac{1}{\frac{1}{2}(\frac{1}{\text{Base}}+\frac{1}{\text{New}})},
\end{equation}
where $\text{Base}$ and $\text{New}$ denote the accuracy of base classes and new classes, respectively.

\textbf{Out-of-Distribution Setting.} In the OOD setting, we initially train the model using the source dataset and then transfer the learned prompts to the target dataset to evaluate the robustness of the method. In our experiments, following COOP, CoCoOp and MaPLe, the source dataset is ImageNet, employing all 1000 of its classes for prompt learning, where each class comprises 16 annotated samples. In the cross-dataset setting, 10 datasets,  Caltech, Pets, Cars, Flowers, Food, Aircraft, SUN, DTD, EuroSAT, and UCF, are target datasets. In the cross-domain setting, the target datasets include ImageNetV2, ImageNet-S, ImageNet-A, and ImageNet-R, all sharing the same classes as ImageNet but differing in distributions.

\section{Additional Experiments}\label{app:exp}
\subsection{Analysis of the Hyper-Parameters}\label{app:exp_params}
\textbf{The number of templates $M$.} Table \ref{tab:ab_m} presents the performance of the model in the base-to-new setting and the corresponding computational complexity when the number of templates changes from 1 to 8. The results in Table \ref{tab:ab_m} are obtained solely using $\mathcal{L}_{t\text{-}ce}$, without the help of \deshort. As the number of templates increases, the performance of the model on the base classes and the new classes increases consistently, while at the same time, the computational overhead increases. Although better performance can be achieved when M is further increased to 8, we select $M=4$ in order to balance the accuracy of classification with the computational overhead.

\begin{table}[!ht]
\caption{Comparison of the performance under different template numbers.}
\centering
\setlength{\tabcolsep}{5.8mm}
\begin{tabular}{c|ccc|cc}
\toprule
M & Base & New & HM & Params & FPS \\ 
\midrule
1 & 80.87  & 73.49  & 77.00 & 3.55M & 600 \\ 
2 & 82.80 & 76.59 & 79.57 & 7.10M & 280.00\\ 
4 & 82.65  & 76.78 & 79.61 & 14.20M & 185.71 \\ 
8 & 83.40  & 77.29  & 80.23 & 28.40M & 65.63\\ 
\bottomrule
\end{tabular}
\label{tab:ab_m}
\end{table}

\begin{figure}[!ht]
\centering
\subfloat[DTD]{\includegraphics[width=.4\columnwidth]{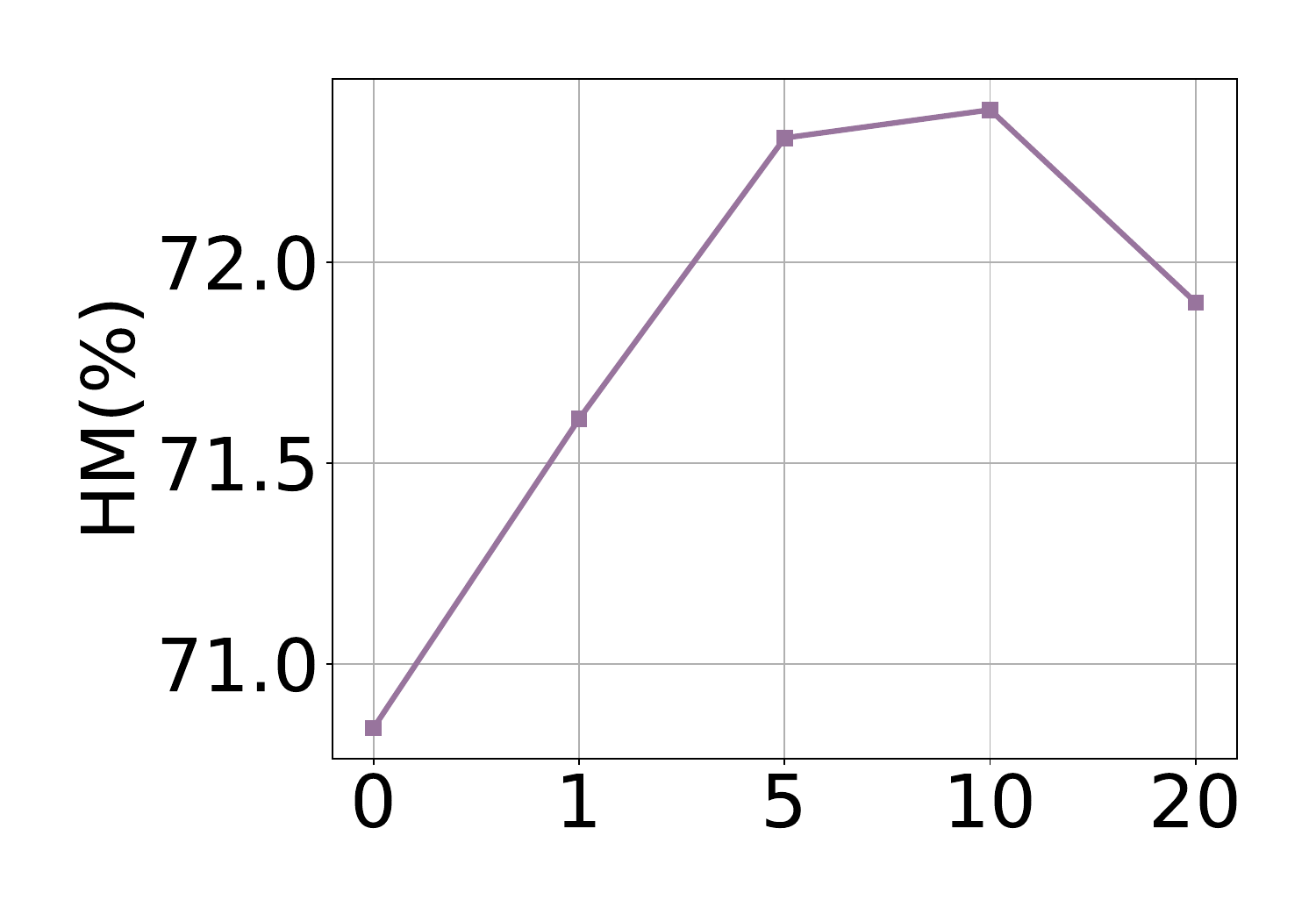}}
\subfloat[FGVC Aircraft]{\includegraphics[width=.4\columnwidth]{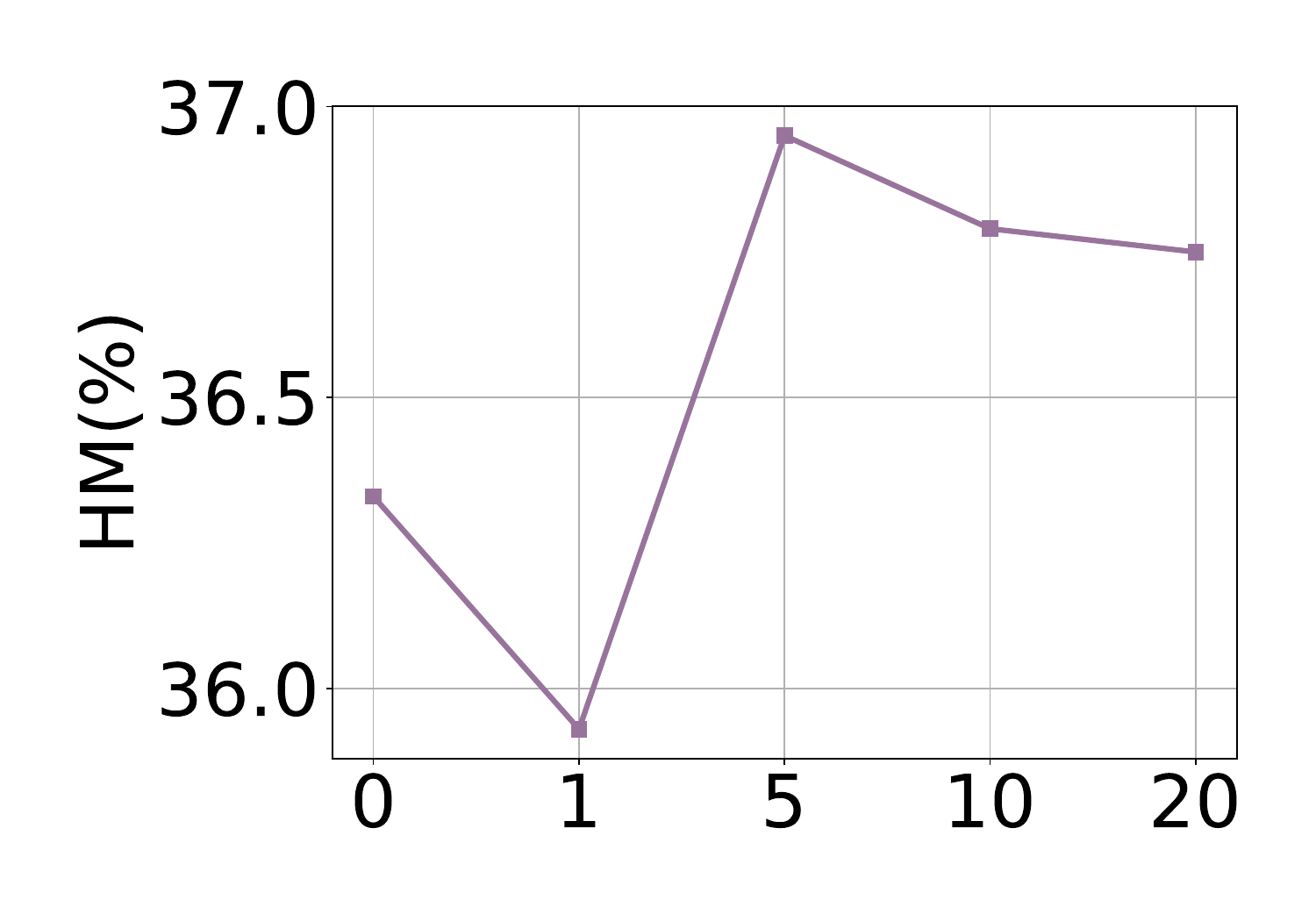}}
\caption{The impact of $\beta$ on performance.}
\label{fig:ab_beta}
\end{figure}

\textbf{Weight $\beta$ for $\mathcal{L}_{de}$. }Figure~\ref{fig:ab_beta} gives the HM on both DTD and Aircraft datasets when varying the weights $\beta$ of $\mathcal{L}_{de}$. From the figure, it is evident that using $\beta=5$ and $\beta=10$ effectively improves the performance of the model on these two datasets as compared to training without $\mathcal{L}_{de}$, i.e., $\beta=0$. When $\beta$ is further increased to 20, the performance of the model begins to decline. Considering the performance on each dataset, in the base-to-new setting, we set $\alpha=5$.

\begin{figure}[!ht]
\centering
\subfloat[UCF101]{\includegraphics[width=.4\columnwidth]{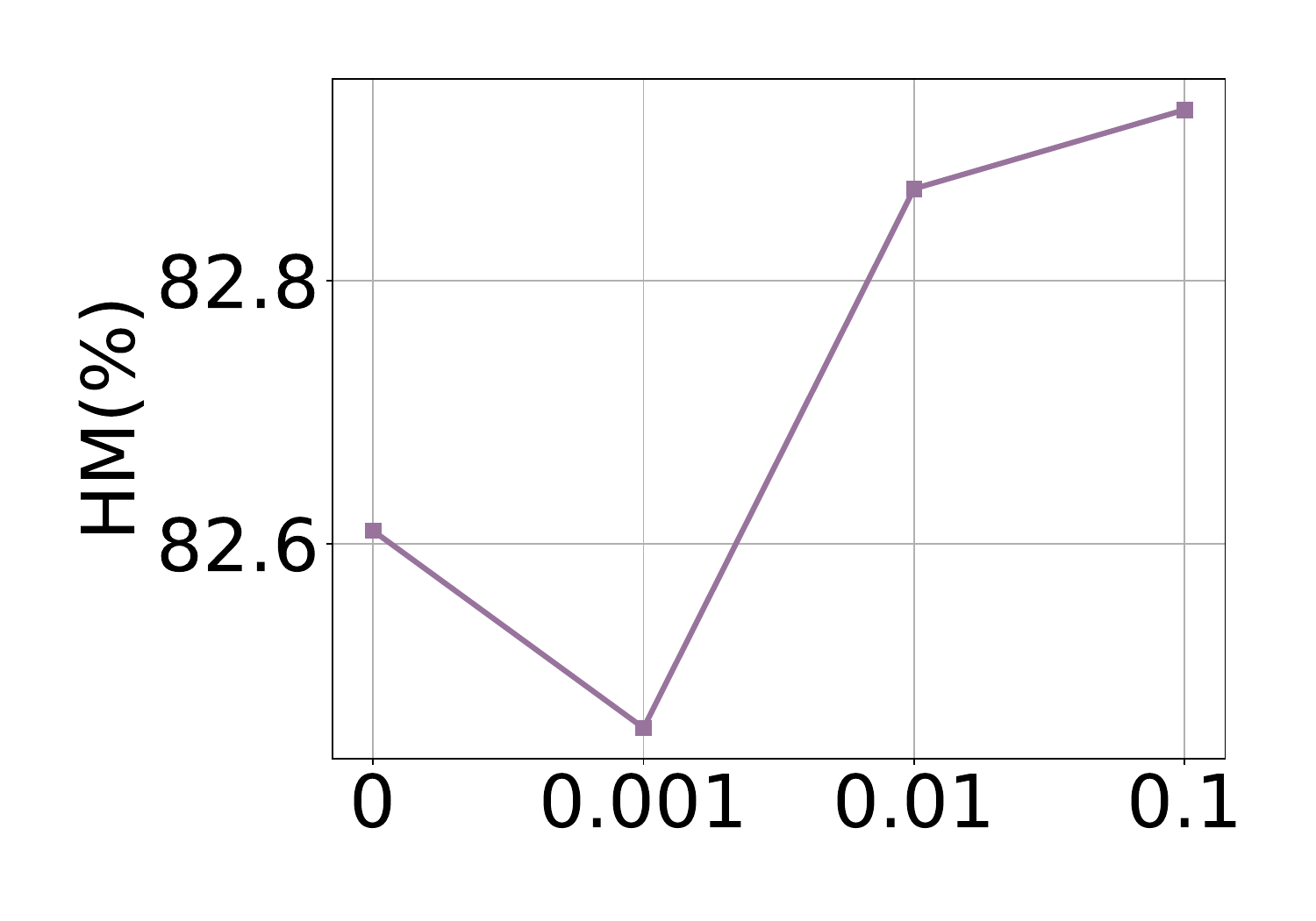}}
\subfloat[Caltech101]{\includegraphics[width=.4\columnwidth]{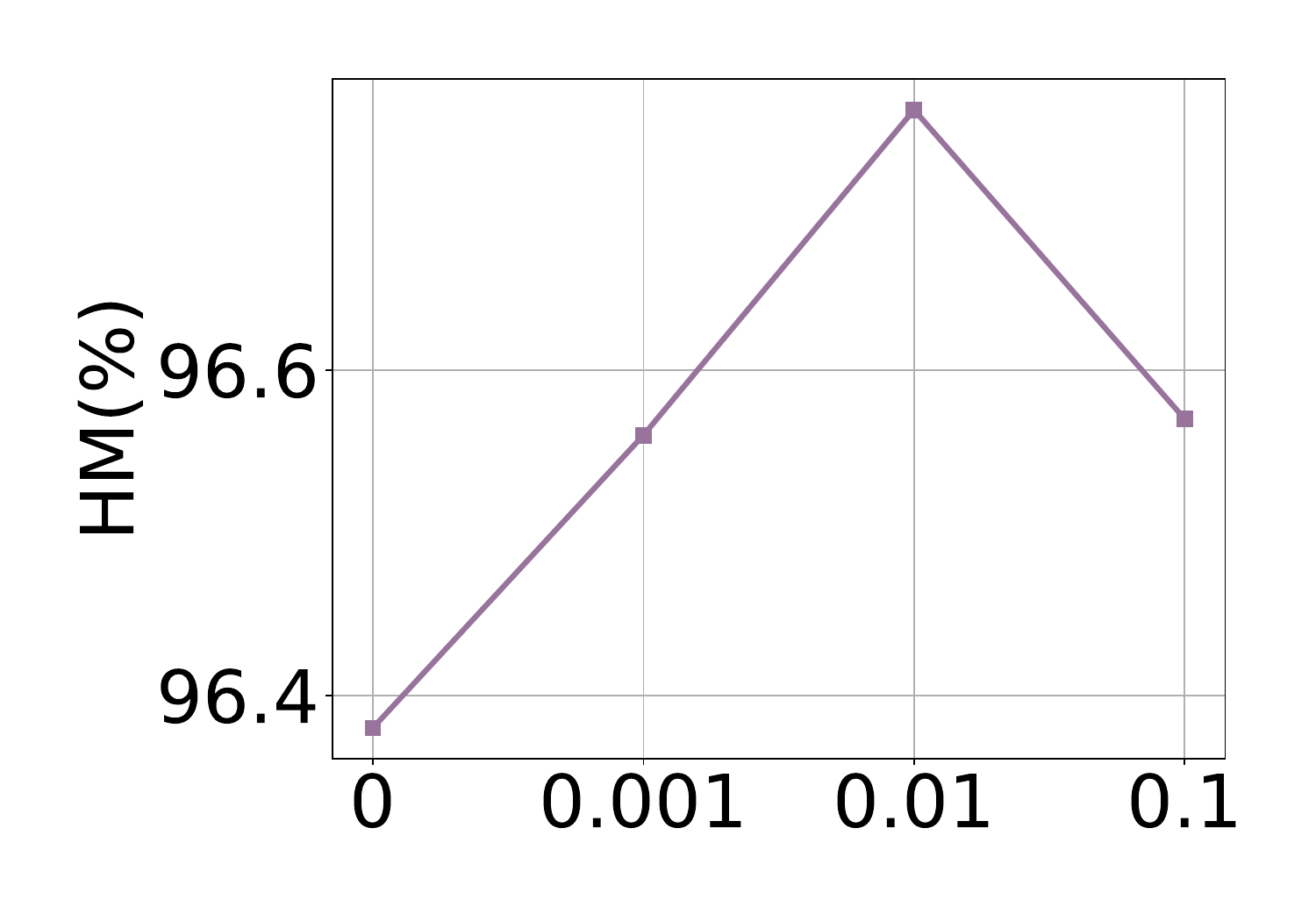}}
\caption{The impact of $\gamma$ on performance.}
\label{fig:ab_gamma}
\end{figure}

\textbf{Weight $\gamma$ for $\mathcal{L}_{con}$. }In Figure~\ref{fig:ab_gamma}, we provide the experimental results obtained by varying $\gamma$ when keeping $\beta=5$. From the results, we observe that setting $\gamma=0.01$ enables the performance of both datasets to obtain an enhancement compared to $\gamma=0.0$. When $\gamma$ is further increased to 0.1, the performance on the Caltech101 dataset shows a significant decrease. Considering all results together, we set $\gamma=0.01$ in all base-to-new experiments.

\textbf{Augmentation methods. }Setting $M=4$, $\beta=5$, $\gamma=0.01$, we vary the augmentation methods to explore the effect of different augmentations on model performance and demonstrate the results in Table~\ref{tab:ab_aug}. We set four sets of augmentations for 4 templates. Specifically, for \textbf{Augmentation 1}, the augmentation methods include: (i) random crop and random flip; (ii) color jitter and random flip; (iii) random translation and random flip; (iv) random augmentation. For \textbf{Augmentation 2}, the four augmentation sets are: (i) random crop and random flip; (ii) random crop, random flip, and color jitter; (iii) random crop, random translation, and random flip; (iv) random crop and other random augmentations. Augmentation 2 adds the random crop to each augmentation set based on Augmentation 1. From Table~\ref{tab:ab_aug}, we notice that for most of the datasets such as ImageNet, Caltech101, etc., there is no great difference in the performance obtained by Augmentation 1 and Augmentation 2. However, for \textit{Cars} and \textit{Aircraft}, Augmentation 2 performs significantly better than Augmentation 1. Capturing discriminative information is challenging for fine-grained classification. We argue that the translation invariance and invariance to the object scale brought by random crop may be critical, so Augmentation 2 performs significantly better than Augmentation1 on these two datasets.

\begin{table}[t]
\caption{The impact of augmentation methods on the performance.}
\centering
\setlength{\tabcolsep}{3.8mm}
\begin{tabular}{l|ccc|ccc}
\toprule
\multirow{2}{*}{Dataset} & \multicolumn{3}{c|}{Augmentation 1} & \multicolumn{3}{c}{Augmentation 2} \\ \cmidrule{2-7} 
& Base & New & HM & Base & New & HM \\ 
\midrule
ImageNet & 77.50 & 71.73 & \textbf{74.51} & 77.30 & 71.57 & 74.32 \\ 
Caltech & 98.63 & 93.67 & 96.09 & 98.20 & 94.37 & \textbf{96.25} \\ 
Pets & 95.97 & 97.67 & 96.81 & 96.07 & 98.00 & \textbf{97.02} \\ 
Cars & 72.33 & 71.63 & 71.98 & 73.80 & 73.97 & \textbf{73.88} \\ 
Flowers & 96.93 & 75.07 & \textbf{84.61} & 96.93 & 74.67 & 84.36 \\ 
Food & 90.93 & 92.23 & 91.58 & 90.87 & 92.33 & \textbf{91.59}  \\ 
Aircraft & 36.83 & 36.23 & 36.53 & 37.47 & 37.50 & \textbf{37.48} \\ 
SUN & 82.37 & 80.03 & \textbf{81.18} & 82.00 & 79.77 & 80.87  \\ 
DTD & 82.70 & 64.10 & \textbf{72.22} & 82.13 & 61.73 & 70.49  \\ 
SAT & 95.10 & 82.33 & \textbf{88.26} & 92.87 & 75.83 & 83.49  \\ 
UCF & 85.70 & 81.73 & \textbf{83.67} & 85.40 & 81.23 & 83.26  \\ 
\bottomrule
\end{tabular}
\label{tab:ab_aug} 
\end{table}

\subsection{Template Study}
In Table~\ref{tab:template}, we provide the full results of our method under all templates, as well as the final performance after performing the evidence fusion which is denoted by \method. From Table~\ref{tab:template}, \method performs the best in 10 out of 11 datasets. Among them, several datasets gain essential improvements compared to the results under a single template, including ImageNet, Flowers, SUN, and DTD. This suggests that the essence of our approach is not to search for better discriminative semantics by increasing the number of templates but to eliminate the impacts of task-independent generative factors by combining multiple sets of decoupled semantics. Despite the moderate performance of any individual template, our method achieves a substantial performance enhancement by fusing them.

\begin{table}[!ht]
\caption{The comparison with the results from different templates.}
\centering
\setlength{\tabcolsep}{5.8mm}
\begin{tabular}{l|cccc|c}
\toprule
Dataset & $S^1$ & $S^2$ & $S^3$ & $S^4$ & \method \\ 
\midrule
ImageNet & 70.10 & 70.10 & 69.73 & 70.33 & \textbf{71.73} \\ 
Caltech & 94.10 & \textbf{94.73}  & 94.57  & 94.17 & 94.37 \\ 
Pets & 97.60 & 97.60 & 97.37 & 97.87 & \textbf{98.00} \\
Cars & 72.93  & 73.57  & 70.87  & 72.63  & \textbf{73.97}  \\ 
Flowers & 72.87  & 73.23  & 74.10  & 71.43  & \textbf{75.07}  \\ 
Food & 91.60  & 91.40  & 91.73  & 91.73  & \textbf{92.33}  \\ 
Aircraft & 34.77  & 35.33  & 35.70  & 35.60 &  \textbf{37.50}  \\ 
SUN & 78.53  & 77.87  & 77.73  & 78.17 &  \textbf{80.03}  \\ 
DTD & 59.60  & 57.83  & 57.87  & 60.43 & \textbf{64.10}  \\ 
SAT & 73.73  & 75.67  & 74.57  & 73.73 & \textbf{82.33}  \\ 
UCF & 77.83  & 76.00  & 77.60  & 78.13 & \textbf{81.73}  \\ 
\bottomrule
\end{tabular}
\label{tab:template}
\end{table}

\subsection{Intrinsic Decoupling of Semantics from Different Templates}
\begin{figure}[t]
\centering
\subfloat[]{\includegraphics[width=0.27\columnwidth]{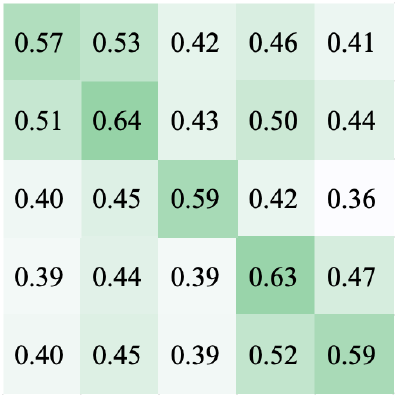}}\hspace{10pt}
\subfloat[]{\includegraphics[width=0.27\columnwidth]{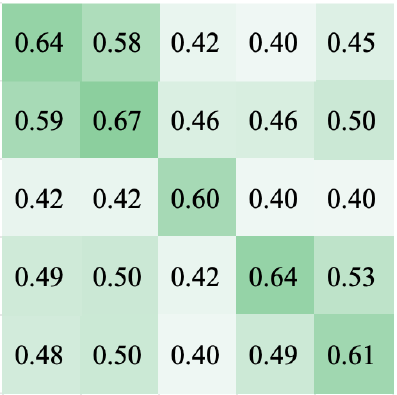}}\hspace{10pt}
\subfloat[]{\includegraphics[width=0.27\columnwidth]{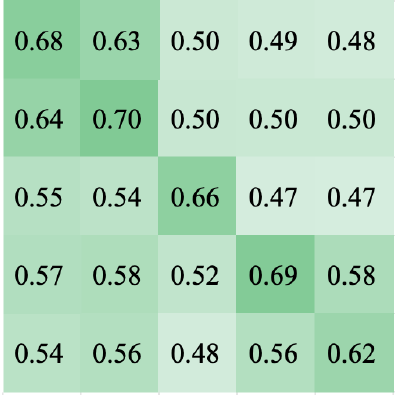}}
\caption{The cosine similarities of the text embeddings under different templates based on the new classes of DTD. (a) The cosine similarities between the text embeddings generated by the 0-th template and the 1-th template. (b) The cosine similarities between the text embeddings generated by the 1-th template and the 2-th template. (c) The cosine similarities between the text embeddings generated by the 2-th template and the 3-th template.}
\label{fig:decouple}
\end{figure}

In \textbf{Section~\ref{sec:ablative}}, we demonstrate that increasing the number of templates can get a performance boost even without explicitly decoupling the semantics contained in different templates. We attribute this to the fact that templates based on different initializations do not learn exactly the same semantics, i.e., the semantics contained in different templates are intrinsically decoupled. To verify this suspicion, we collect the text embeddings obtained based on different templates without applying \deshort, and compute the similarities between the embedding obtained from different templates. We plot the results of the DTD dataset in Figure~\ref{fig:decouple}. As can be seen from the figure, even for the same category, the similarities between the text embedding obtained by different templates are still much less than 1 (i.e., the value on the diagonal line in the figure). This indicates that the text embedding obtained by different templates has a large discrepancy, which confirms our idea that the semantics learned by different templates are inherently decoupled. Therefore, better performance can be obtained through fusing results generated by these decoupled features.

\end{document}